\documentclass{article}

\usepackage{arxiv}

\usepackage[utf8]{inputenc} 
\usepackage[T1]{fontenc}    
\usepackage{hyperref}       
\usepackage{url}            
\usepackage{booktabs}       
\usepackage{amsfonts}       
\usepackage{nicefrac}       
\usepackage{microtype}      
\usepackage{lipsum}		
\usepackage{graphicx}
\usepackage{natbib}
\usepackage{doi}

\usepackage{CJKutf8}
\usepackage{xcolor}
\usepackage{amsmath,amsfonts}
\usepackage{algorithm}%
\usepackage{algorithmicx}%
\usepackage{algpseudocode}%
\usepackage{array}
\usepackage[caption=false,font=normalsize,labelfont=sf,textfont=sf]{subfig}
\usepackage{textcomp}
\usepackage{stfloats}
\usepackage{url}
\usepackage{verbatim}
\usepackage{balance}

\usepackage{algpseudocode}%
\usepackage{listings}%
\usepackage{subcaption}
\usepackage{multirow}
\usepackage{booktabs} 
\usepackage{adjustbox}
\usepackage{dirtytalk}

\usepackage{rotating}
\usepackage{makecell}
\usepackage{booktabs}
\usepackage{array}

\title{Maswao: An automatically collected, prepared, and benchmarked dataset for South African Sign Language}

\author{ \href{https://orcid.org/0000-0002-9754-9547}{\includegraphics[scale=0.06]{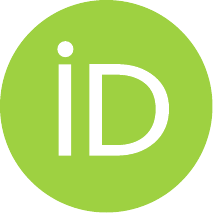}\hspace{1mm}Thapelo A.~Sindane} \\
	Data Science For Social Impact Lab\\
    Department of Computer Science\\
	University of Pretoria\\
	\texttt{sindane.thapelo@tuks.co.za} \\
	\And
	  Unarine L.~Netshifhefhe \\
	Data Science For Social Impact Lab\\
    Department of Computer Science\\
	University of Pretoria\\
    \texttt{u20505656@tuks.co.za} \\
	\AND
	  Vukosi Marivate \\
	Data Science For Social Impact Lab\\
    Department of Computer Science\\
	University of Pretoria\\
    Lelapa AI \\
	\texttt{vukosi.marivate@cs.up.ac.za} \\
	\And
	Abiodun Modupe \\
	Data Science For Social Impact Lab\\
    Department of Computer Science\\
	University of Pretoria\\
	\texttt{abiodun.modupe@cs.up.ac.za} \\
	\And
	Herkulaas M.~Combrink \\
	interdisciplinary center for digital futures \\
    Faculty of Economic and Management Sciences \\
	University of the Free State \\
	\texttt{CombrinkHM@ufs.ac.za} \\
    \And
	Ketro Sithole \\
	Data Science For Social Impact Lab\\
    Department of Computer Science\\
	University of Pretoria\\
	\texttt{u18312889@tuks.co.za} \\
}

\renewcommand{\shorttitle}{\textit{arXiv} Template}

\hypersetup{
pdftitle={A template for the arxiv style},
pdfsubject={q-bio.NC, q-bio.QM},
pdfauthor={David S.~Hippocampus, Elias D.~Striatum},
pdfkeywords={First keyword, Second keyword, More},
}

\begin{document}
\maketitle

\begin{abstract}
	Recent advancements in sign language processing can be attributed to developments in both large-scale dataset collection and algorithmic complexity in areas such as pose estimation techniques, signal processing through Transformer models, and model optimization. However, only a handful of sign languages, such as the American, British, and German, from over 200 globally recognized sign languages, reap the benefits of such advancements, while the rest lag behind. In this study, we aim to address this technological disparity by proposing a framework for the automatic collection and preparation of sign language datasets, with very minimal human-labour intervention, using the South African Sign Language (SASL) as a testbed. Specifically, we automatically collect SASL video content using public resources such as news broadcasts. Additionally, we convert their accompanying audio streams into text using available Automatic Speech Recognition (ASR) tools. Furthermore, we crop the sign language area in the videos and use gradient descent to locate valleys of non-motion to determine end-of-sentence points within long videos to create our weakly sentence-aligned parallel corpora -- the Maswao (Signs) dataset. Finally, we benchmark the weakly aligned dataset on the task of Sign-to-text Translation using Recurrent Neural Networks and Transformer models to compare their individual intricacies. Our results show that Transformer models proposed in the literature outperform recurrent baselines, with the best-performing Transformer achieving BLEU-1 to BLEU-4 scores of 13.78, 5.21, 2.52, and 1.54, respectively, as well as ChrF = 18.09 and ROUGE-L = 10.51. These results indicate that the task remains promising and warrants further investigations.
\end{abstract}


\section{Introduction}\label{sec1}
Sign-to-text translation (STT) and Text-to-sign translation (TTS) are two complementary tasks aimed at enabling collegial communication between the Deaf and hearing communities~\cite{hwang2025spatio}. In many cases, these two important subcomponents are addressed independently, with STT spearheading the race. 

STT evolved from the task of sign language recognition~\cite{kong2014towards}, and can be defined as the process of translating complete sign-language sentences into grammatically sound spoken language sentences~\cite{camgoz2018neural}. Many works have been proposed to tackle this challenge, with earlier methods constrained by feature extraction techniques~\cite{tharwat2015sift,zhang2016multi} transitioning to more advanced methods using convolutional neural networks (CNN), due to their enhanced hierarchical feature extraction  mechanisms~\cite{tran2015learning}. 

More recently, prominent STT techniques adopted hybrid architectures comprised of deep CNN feature extractors such as ResNet~\cite{deng2009imagenet}, I3D~\cite{carreira2017quo}, and Inception~\cite{szegedy2015going} to model spatial relationships within the input images, and Transformer models~\cite{devlin2019bert} to learn the temporal dependencies of input sequences, drawing from advancements in Natural Language Processing. Regardless, the quality of available video datasets (e.g., poor video resolutions, inconsistent signer positioning, variation in signs, poor illumination, etc), and the inaccuracy of pose estimation techniques remain as the main challenges in STT to date~\cite{de2020sign}.  

Text-to-sign (TTS) on the other hand is the reverse process of STT~\cite{zhao2000machine}. Historically, TTS has often been approached as an avatar-rendering problem~\cite{cox2002tessa}. However, due to a myriad of issues such as avatar stiffness, unnatural motion, the complexity of converting input signals into parametric representations, increased annotations costs, and high reliance on expert knowledge, research alternatives such as deep learning models using recurrent neural networks~\cite{stoll2018sign} and recently Transformer models~\cite{saunders2020progressive}, Diffusion models~\cite{tang2025gloss}, and generative adversarial neural networks~\cite{saundersadversarial} that generates photo-realistic sequences of sign sequences, are gaining momentum. Despite the impressive performance reported by these techniques, several challenges remain. These include regression to the mean~\cite{wahbi2023semi}, the accumulation of marginal errors (often referred to as drift)~\cite{saunders2020progressive}, and poor perceptual responses from Deaf communities~\cite{ventura2020can}, which continue to limit practical adoption and impact.

\begin{figure*}[!t]
\centering
\subfloat[]{\includegraphics[width=2.5in]{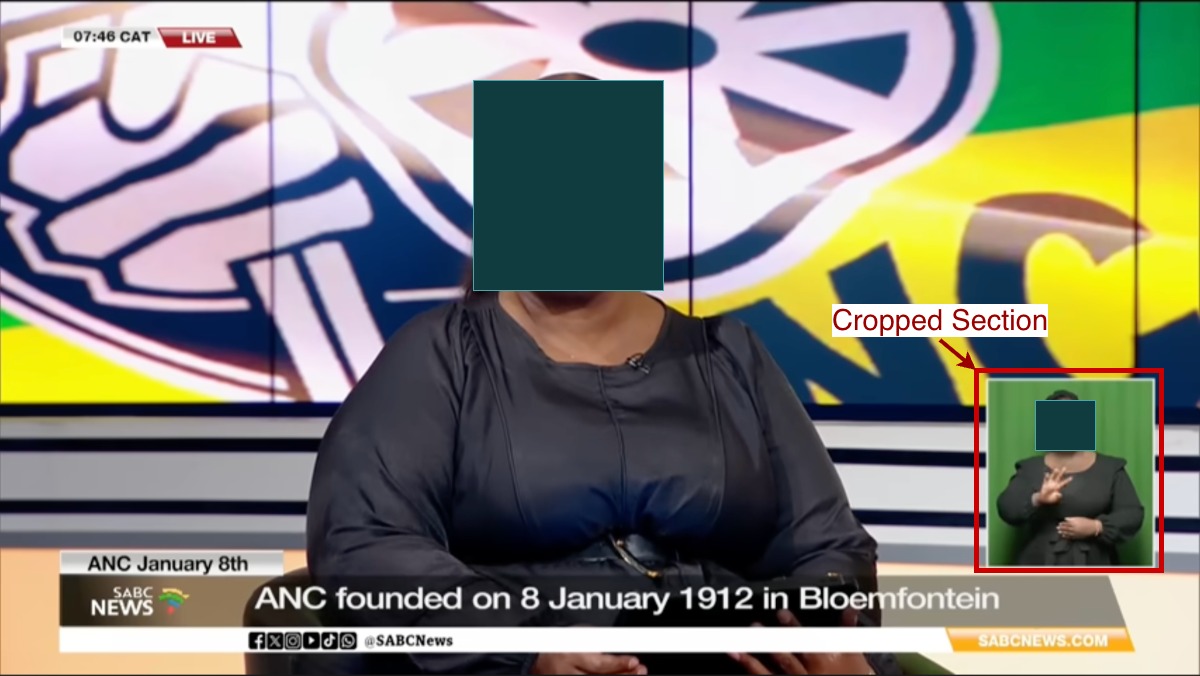}%
\label{fig:sabc_example}}
\hfil
\subfloat[]{\includegraphics[width=2.5in]{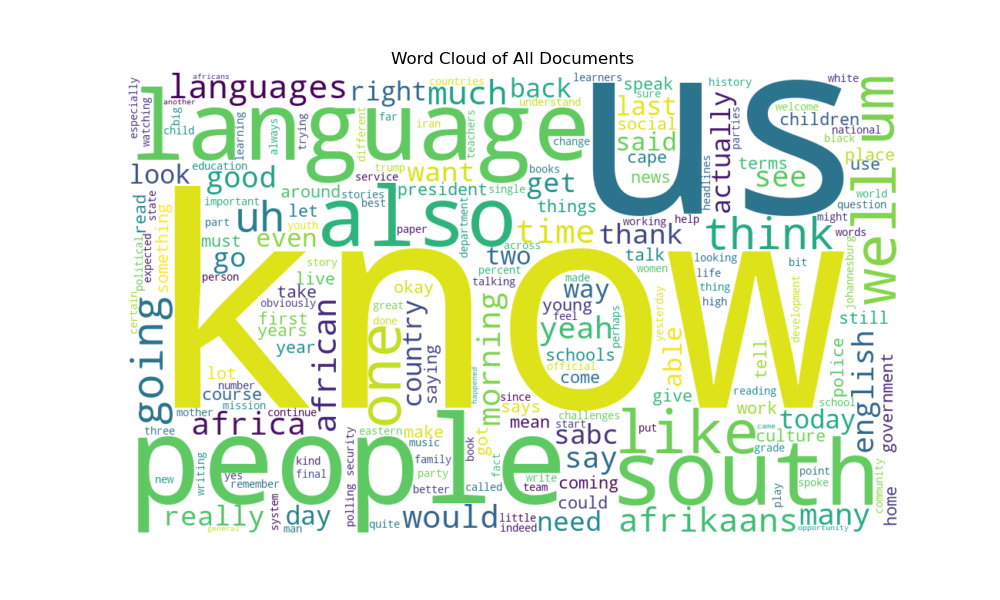}%
\label{fig:topics}}
\caption{News video example with sign language interpretation example (left) with topic model for all collected videos (right).}
\label{fig_sim}
\end{figure*}

Importantly, the fuel that drives these rapid developments in both STT and TTS is the large-scale collection of parallel, annotated, and quality sign language datasets. The availability, and further refinement of such datasets remains the primary drivers of success in sign language processing. For example, datasets such as the PHOENIX2014\textbf{T}~\cite{koller2015continuous}, Chinese sign language dataset (CSL)~\cite{pu2020boosting}, and the How2Sign~\cite{duarte2021how2sign} are among the highly evaluated non-proprietary datasets in sign language processing history, with recent algorithmic advancements such as Transformers benchmarked, their limitations scribed, and state-of-the-arts (SOTA) established on them ~\cite{fang2024signllm}. However, similar benchmarks that shape research directions have not been established for many low-resourced sign languages, including South African Sign Language (SASL). 

To address this gap, we propose a framework for the automatic collection, preparation, and benchmarking of SASL data that aims to minimise the resources required for dataset development while maintaining data quality. The framework is designed to be extensible to other low-resource sign languages with similar resource characteristics, such as the availability of unstructured public sign-language video content. More specifically, our collection includes the crawling of YouTube videos from a popular and notarized South African news outlet - the South African Broadcasting Corporation (commonly known as the SABC), which occasionally releases news video broadcasts that include a sign language interpreter, translating the presenter’s spoken utterances into SASL as shown in Figure~\ref{fig:sabc_example}~\cite{SABC}.  

From this source, we created the first weakly aligned spoken language to sign language dataset with over 2944 text-video sentence-level examples, covering a wide vocabulary (see Figure \ref{fig:topics}). Lastly, to accelerate research in STT for SASL, we benchmark the dataset using a set of state-of-the-art techniques, including MMPose~\cite{MMPose2020} for pose estimation, AlexNet-based CNN feature extraction, Recurrent Neural Networks, and Transformer models. This benchmarking establishes an initial state-of-the-art baseline for Maswao SASL dataset using an AlexNet + Transformer architecture. 

The contribution of this paper is succinctly organized as follows:
\begin{enumerate}
    \item We present the first weakly aligned parallel video-to-text corpus for South African Sign language (SASL) using publicly available news content - the Maswao (meaning 'signs' in Northern Sesotho) dataset. This dataset comprises 2,944 weakly aligned spoken-language sentences paired with their equivalent sign-language video sentence representations.
    \item We present the first benchmark on STT, using various pose estimation tools, deep CNNs, Recurrent networks, and Transformer models. 
    \item We establish a state-of-the-art AlexNet+Transformer model for our weakly aligned Maswao SASL dataset on the STT task.
\end{enumerate}

Our work aims to accelerate advancements in South African Sign Language (SASL) research by proposing a reproducible and low-cost framework for the automatic curation of sign language datasets. To foster open and reproducible research, we publicly release our code, the Maswao dataset, and supporting resources here~\footnote{https://github.com/dsfsi/maswo-south-african-sign-language-dataset.git}, enabling other researchers to build upon and extend this work.

\section{Related Works}
According to the World Health Organization (WHO), there are over 466 million people suffering from some form of hearing loss, ranging from mild to severe hearing deficiency~\cite{whowebsite,whoWorldReport}. The WHO further projected that by the year 2070, approximately 1 billion people of the global population will be affected by hearing loss~\cite{kushalnagar2019deafness}. The primary mode of communication for many of these hearing-impaired individuals is sign language, with a footprint of over 200 diverse languages, with their own developed and complex grammar and structure. However, while communication between Deaf people with knowledge of the locally used sign language is possible, communication between the majority of the Deaf and the hearing communities presents many barriers reppling down to access challenges (e.g., health facilities, security services, entertainment, and many more). Interpreters have mostly been used to mediate communication between the Deaf and the hearing communities; however, the limited number of qualified interpreters often fail to meet the high demand~\cite{mohamed2022deep,rastgoo2022all}. Often leading to vulnerabilities in society~\cite{sanda2014aftermath}. Consequently, automated sign-to-text translation and text-to-sign are emerging as promising approaches for bridging the communication divide.

End-to-end Sign-to-text (STT) translation systems historically adopted statistical approaches such as Hidden Markov Models~\cite{dreuw2007speech,zafrulla2011american}. These methods relied on hand-engineered features, limiting the quality and scalability of their approach to domain-constrained conditions. Subsequent to statistical approaches, classical machine learning techniques such as Conditional Random Fields, Random Forest, Support Vector Machines, and later multi-layered perceptrons took form~\cite{kong2014towards}. However, these methods suffered the same fate as their predecessors as they also relied on feature extraction techniques to extract discriminative features~\cite{yang2008sign,chavan2016indian}. Recently, with the advent of sophisticated deep neural network architectures, many STT pipelines have adopted complex systems using deep convolutional neural networks such as ResNet, EfficientNet, I3D, and Inception as their backbone, due to their capabilities in automatically extracting abstract latent representations of spatial data~\cite{alharthi2023vision}. Others have adopted deep recurrent neural networks such as LSTMs and GRUs, due to their ability to efficiently model temporal dependencies between words~\cite{camgoz2018neural}. However, a new wave of Transformer models drawing from successes in NLP has marked a new paradigm shift~\cite{ko2019neural, ananthanarayana2021dynamic, singla2024enhancing, hu2023signbert+, sreemathy2024indian}, eliciting magnitudes of performance gains from the former methods, as datasets continue to become complex and large. For example, \cite{de2020sign} compares LSTM models with Transformer models, using features extracted using OpenPose and ResNet-43. Their study shows that, while LSTM models perform better on the development samples, Transformer models towered over them on unseen instances, highlighting their generalization capabilities. Similarly, \cite{camgoz2020sign} proposed the first implementation of a Transformer architecture on the challenging PHOENIX2014\textbf{T} dataset, establishing a new state-of-the-art succeeding recurrent predecessors. \cite{yin2020sign} improved on the state-of-the-art by proposing a spatio-temporal multi-cue model stacked with a Transformer encoder for STT. \cite{de2021frozen} leverages the popular transfer learning technique using pre-trained BERT \cite{devlin2019bert}, and mBART-50 \cite{tang2020multilingual} models to translate input videos into spoken language sequences. Recently, \cite{lopez2025multitask} presents a CNN-based feature extractor using motion primitives derived from optical flows as inputs, which are then passed into a Transformer model for STT translation. Their comprehensive evaluation on multiple datasets: PHOENIX2014\textbf{T}, and Colombian sign language (CoL-SLTD) datasets, showed the effectiveness of their proposed technique.

Text-to-sign translation systems on the other hand adopted avatar rendering applications to produce sign language utterances from spoken language inputs such as text or voice~\cite{zhao2000machine,cox2002tessa,cox2003development,lincoln2003comparison,glauert2006vanessa}. However, issues such as restricted domain of discource leading to low vocabularies, increased annotation cost, and high reliance on specialized skillset for extracting motion capture, limited the application of these methods. To address these shortcomings, subsequent methods adopted deep learning methods such as recurrent neural networks, coupled with the emergence of pose estimation tools such as OpenPose~\cite{cao2019openpose}, and MediaPipe~\cite{lugaresi2019mediapipe}, to predict a sequence of joint coordinates of skeletal sequences $P = \{ P^{K}_{M}\}$ with $M$ frames and $K$ joints, equivalent to sign language movements, given a spoken language sequence $X = \{x_{1}, x_{2}, ..., x_{W} \}$, with $W$ words~\cite{stoll2020text2sign}. Recurrent neural networks, and alternate modifications such as LSTMs, Gated Recurrent Units, etc, are notoriously susceptible to the vanishing gradient problem~\cite{hochreiter1998vanishing}, and insufficiency in modeling long-temporal dependencies. Although hybrid models based on recurrent networks and attention mechanisms were proposed, they were bounded by their costly computational requirements, stemming from recurrent processing~\cite{stoll2018sign}. Following these methods, Transformer-based approaches, due to enhanced contextual representations, gained interest \cite{zelinka2020neural,saunders2020progressive}. Consequently, \cite{saundersadversarial} proposed a Transformer-based generative adversarial network that receives spoken language input sentences and converts them into pose-based sequences congruent to sign language gestures. \cite{saunders2021mixed} on the other hand proposed a system of multiple Transformer models, each specializing in specific feature representations, trained jointly to optimize the collective goal of sign language translation. \cite{cui2022pyramid} presents a non-auto-regressive decoding scheme as a solution to mitigating the regression to the mean challenges, resulting in an interesting line of work building on their foundation~\cite{azevedo2024empowering,ri2024tamil}. \cite{fang2024signllm} proposes a large language model (LLM) framework that trains multiple Transformer models across several spoken-language–to–sign-language pairs. Meanwhile, diffusion-based and Transformer-based approaches are gaining traction for translating input text sequences into pose-based skeletal representations~\cite{tang2025gloss,wang2025linguistics}. 

The continued advancement of both STT and TTS research relies heavily on the availability of large-scale, high-quality, annotated datasets. In this regard, \cite{duarte2021how2sign} introduced How2Sign, a multimodal and multi-view American Sign Language (ASL) corpus comprising over 80 hours of recordings paired with corresponding gloss annotations for sign language translation. Despite its scale, the dataset is limited to a single sign language and primarily supports translation between ASL and English. More recently, \cite{tanzer2025youtube} leveraged YouTube to construct a large-scale, open-domain multilingual sign language dataset spanning 25 sign languages, with the primary objective of translating sign language into the corresponding regional spoken language. Similarly, \cite{avramidis2025tub} proposed a multilingual sign language corpus covering 12 sign languages, collected from publicly available sources including news broadcasts, government communications, and educational content. Although these datasets have substantially advanced sign language research, they include little to no representation of African sign languages or African spoken languages, highlighting a significant gap in the available resources for the African context.
   
South African Sign language is amongst the many under-represented languages, with only four findable literature works proposing technological solutions for sign language processing~\cite{van2006south,mcinnes2014south,ndlovu2022application,chakoma2023converting}, spanning over 3 decades. Specifically, \cite{van2006south} presented a Synchronous Tree-Adjoining Grammar (STAG) parser that transforms input sentences into parse trees used to parametrize an avatar generation model. \cite{mcinnes2014south} subsequently developed a low-cost hand glove for sign gesture collection. \cite{ndlovu2022application} uses smartphone cameras to collect their sign language dataset, and evaluate the quality of their processed datasets using classical feedforward neural networks. Recently, \cite{chakoma2023converting} presents a look-up approach that attempts to match input gestures with pre-stored sign gestures for TTS translation. However, these existing efforts focus on hand gestures or isolated signs and do not extend to sentence-level sign language translations, a limitation that the Maswao dataset seeks to address. Notably, none of the successive works from the four aforementioned work builds on the progress of existing work, rendering literature in the South African context extremely fragmented. Additionally, the two recent works \cite{ndlovu2022application,chakoma2023converting}, do not explore Transformer advancements. To address these issues and more, we propose a continuous framework that includes the collection, and preparation of digital video resources from publicly accessible news outlets, which are then used to benchmark various processing techniques such as pose estimation tools, recurrent neural networks, and Transformer architectures.

\section{Methodology}\label{methodology}
\begin{figure*}[t]
    \centering
    \includegraphics[
        width=\textwidth,
        height=0.9\textheight,
        keepaspectratio
    ]{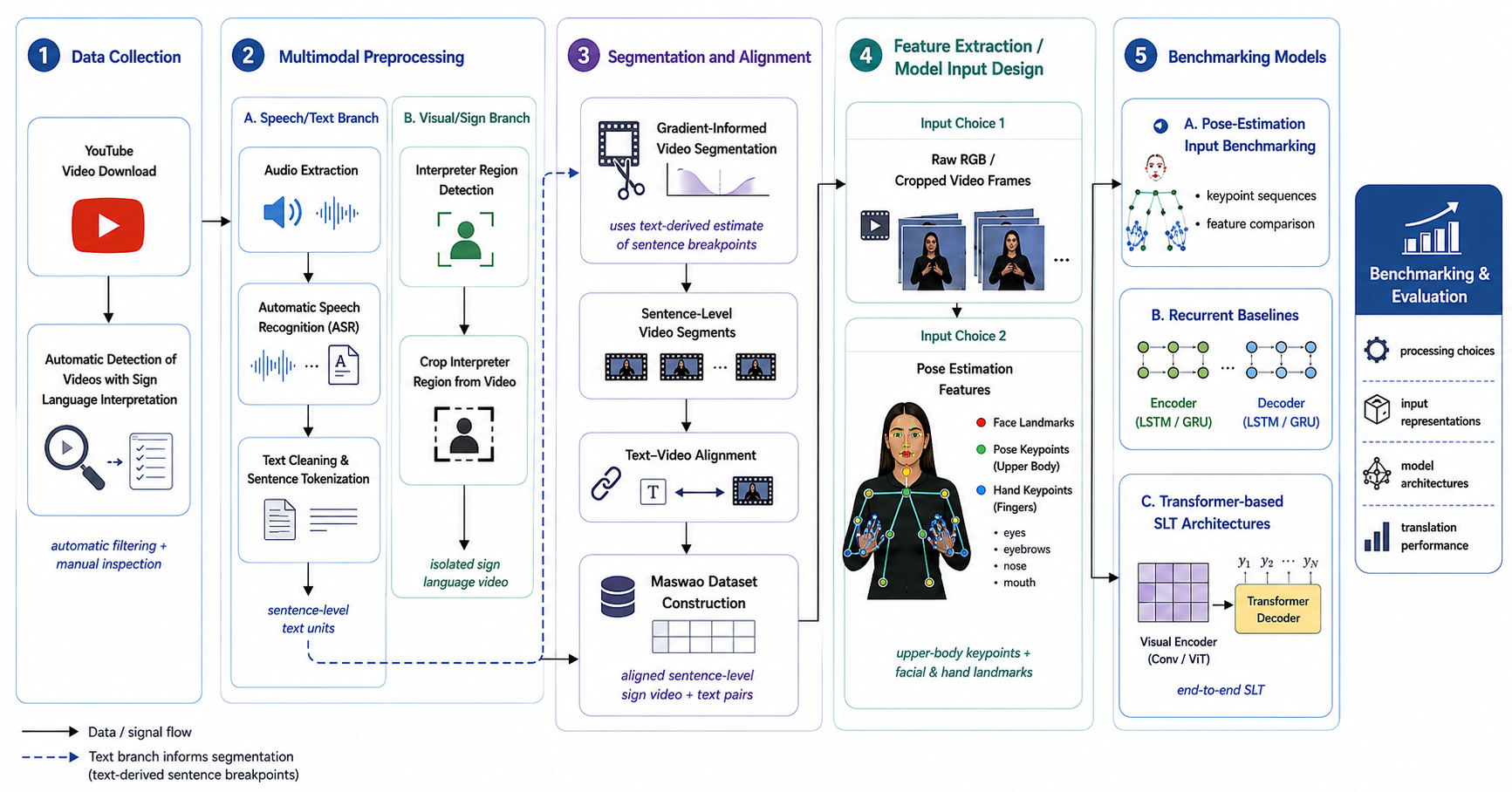}
    \caption{Overview of the framework used to create the Maswao dataset.}
    \label{fig:maswao_framework}
\end{figure*}

In this section we present a largely automated end-to-end pipeline to build a weakly aligned sign-to-text (STT) data resource from publicly accessible news broadcasts. Figure \ref{fig:maswao_framework} shows a summary of our framework including: (i) downloading Youtube video created by SABC for all channels (ii) identifying and filtering videos without sign language interpretation (iii) generating transcriptions from the accompanying audio streams using ASR (iv) cropping and isolating the interpreter section of the videos and (v) deriving sentence-level video-to-text pairs via our proposed motion and gradient-based segmentation technique. This process resulted in our Maswao weakly aligned corpora. We then use the resulting Maswao corpus to benchmark key processing and modeling choices, including pose-estimation inputs, recurrent baselines, and Transformer-based SLT architectures.

\subsection{Data Creation Processes}
Digital sign language content is difficult to find for many low-resource sign languages. In the South African context, the South African Broadcasting Corporation (SABC) is one of the popular public news outlets that occasionally delivers spoken language news content with parallel sign language interpretation. The interpretation is done using the official SASL. The SABC releases their news recording on YouTube, where we crawled news feeds for all 11 official South African languages. To crawl the datasets, we used a python open-source library yt-dlp~\footnote{https://pypi.org/project/yt-dlp/} to download Youtube videos, with properties such as channel (highlighting the language) set as ( English, Sepedi, Sesotho, Setswana, IsiZulu, etc), and extension type $.webm$ to download videos of high quality where options are available -- alternatively, we resorted to $.mp4$. Our capacity to download videos at a larger scale was restricted by Youtube bot detectors, requiring constant authentication after a few downloads, which we managed using manually downloaded cookies files. As such, we were able to download a total of only $2000$ videos from SABC for all channels combined, together with their audio streams. In this study, we only report the implementation of the framework and experiments for the English (en) language, as the language with the largest crawl, and categorized the other official languages as the \say{Others} as shown in Table \ref{tab:Maswao_statistics} describing the statistics of our Maswao dataset. As a result, the initial crawl for English resulted in a raw dataset of $377$ raw videos. However, not all videos downloaded included sign language interpretation. Which meant that we had to first identify videos with interpreters included from the downloaded set, locate the position of the interpreter, and then crop the section with sign language interpretations. Fortunately, the SABC has a standard for placing a sign language interpreter in a box at the bottom right corner of the video frame, solving the location challenge, as shown in Figure \ref{fig:sabc_example}. This reduces the search space for interpreter-box location. 

\begin{table*}[t]
\centering
\small
\setlength{\tabcolsep}{2pt}
\renewcommand{\arraystretch}{1.15}
\begin{adjustbox}{max width=\textwidth}
\begin{tabular}{c|ccccccccccccc|cc|ccc}
\toprule
& \multicolumn{13}{c|}{\textbf{Video Statistics}} & \multicolumn{2}{c|}{\textbf{Extremes}} & \multicolumn{3}{c}{\textbf{Text Statistics}} \\
\cmidrule(lr){2-14} \cmidrule(lr){15-16} \cmidrule(lr){17-19}
Lang. &
\rotatebox{90}{\makecell{Num.\\Raw\\Videos}} &
\rotatebox{90}{\makecell{Videos\\w/\\Sign}} &
\rotatebox{90}{\makecell{Avg.\\Resolution}} &
\rotatebox{90}{\makecell{Std.\\Resolution}} &
\rotatebox{90}{\makecell{Var.\\Resolution}} &
\rotatebox{90}{\makecell{Avg.\\FPS}} &
\rotatebox{90}{\makecell{Std.\\FPS}} &
\rotatebox{90}{\makecell{Var.\\FPS}} &
\rotatebox{90}{\makecell{Total\\Hours}} &
\rotatebox{90}{\makecell{Avg.\\Video\\Length}} &
\rotatebox{90}{\makecell{Std.\\Video\\Length}} &
\rotatebox{90}{\makecell{Var.\\Video\\Length}} &
\rotatebox{90}{\makecell{No.\\Signers}} &
\rotatebox{90}{\makecell{Largest\\Video\\(len.)}} &
\rotatebox{90}{\makecell{Smallest\\Video\\(len.)}} &
\rotatebox{90}{\makecell{Num.\\Sent}} &
\rotatebox{90}{\makecell{Num.\\Words}} &
\rotatebox{90}{\makecell{Unique\\Vocab}} \\
\midrule
en &
377 &
67 &
1511 &
565 &
$3.19\times10^{5}$ &
28.66 &
2.22 &
4.91 &
5.72 &
5.12 &
4.55 &
20.68 &
--    &
20.73 &
0.84 &
2944 &
50986 &
7815 \\

Others &
1623 & -- & -- & -- & -- & -- & -- & -- & -- & -- & -- & -- & -- & -- & -- & -- & -- & -- \\
\bottomrule
\end{tabular}
\end{adjustbox}
\caption{Represents the statistics of the video and text datasets. Shortened words include Number (Num), with (w/), Average (Avg.), Standard deviation (Std), Variance (Var), Frames per second (FPS), Sentence (Sent), and Vocabulary (vocab). The unit of measure for resolution is pixel. The Avg., Std., and Var. video length, and the Largest/Smallest Video columns, are measured in minutes. Largest video: aerttydScic (20m44s). Smallest video: lmMkRta0Se0 (0m50s).}
\label{tab:Maswao_statistics}
\end{table*}

Following this, our automated interpreter box isolation pipeline first extracts video frames from each video at predefined timestamps of 35 seconds and 65 seconds chosen empirically. Each frame is then divided into four equal quadrants by placing one horizontal line and one vertical line through the centre of the frame. Then, MediaPipe Face Detection~\cite{lugaresi2019mediapipe} is applied to the bottom-right quarter regions of these frames using a confidence threshold of 0.3, where the returned normalized face coordinates are converted into pixel values and adjusted back to the original frame coordinates to establish the interpreter’s face position. We focus on this region as this is where the sign language interpreter is expected to appear, reducing unnecessary detections from the main video content. If a face is detected, the pipeline proceeds to the next steps, otherwise a video is skipped for a manual checking reducing the number of videos manually checked. For videos detected to contain a face, the final interpreter box is generated by expanding the detected face bounding box using padding parameters ("1.8 × face\_width" horizontally, "0.5 × face\_height" above, and "3.5 × face\_height" below), ensuring that the crop includes the interpreter’s upper body and surrounding white-bordered area of the interpreter box. The quality of videos, and actual size of the interpreter box varied across videos, making automatic box detection with static bounding box extension sizes sometimes challenging. Therefore, we manually checked failure cases of the automatic extraction method and manually cropped the interpreter videos using the Mediapipe interface to isolate the interpreter section from the rest of the video. 

\begin{table*}[t]
\centering
\small
\label{tab:interpreter_methods_evaluation}
\resizebox{\textwidth}{!}{%
\begin{tabular}{lrrrrrrrrrrr}
\toprule
Method & Total & Pred. & GT & TP & FP & FN & TN & Prec. & Rec. & F1 & FPR \\
\midrule
Maswao-Interpreter-Extractor & 377 & 82 & 67 & 55 & 27 & 12 & 283 & 0.671 & 0.821 & 0.738 & 0.087 \\
\bottomrule
\end{tabular}}
\caption{Performance of the MediaPipe-based interpreter detection on the English subset, evaluated against the manually labelled ground truth.
\textbf{Total}: number of English videos processed.
\textbf{Pred.}: videos the method flagged as containing an interpreter.
\textbf{GT}: videos that contain an interpreter according to the manually corrected ground truth.
\textbf{TP} (true positives): flagged by the method and present in the ground truth.
\textbf{FP} (false positives): flagged by the method but absent from the ground truth.
\textbf{FN} (false negatives): not flagged by the method but present in the ground truth.
\textbf{TN} (true negatives): not flagged and absent from the ground truth.
\textbf{Prec.}: $TP/(TP+FP)$.
\textbf{Rec.}: $TP/(TP+FN)$.
\textbf{F1}: harmonic mean of precision and recall.
\textbf{FPR}: false positive rate, $FP/(FP+TN)$.}
\label{interpreter_methos_evaluation}
\end{table*}

\begin{figure*}[!t]
\centering
\subfloat[Frame 1]{\includegraphics[width=0.16\textwidth]{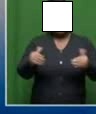}%
\label{fig:frame1}}
\hfil
\subfloat[Frame 2]{\includegraphics[width=0.16\textwidth]{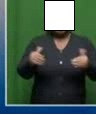}%
\label{fig:frame2}}
\hfil
\subfloat[Frame 3]{\includegraphics[width=0.16\textwidth]{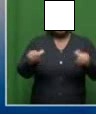}%
\label{fig:frame3}}
\hfil
\subfloat[Frame 4]{\includegraphics[width=0.16\textwidth]{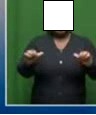}%
\label{fig:frame4}}
\hfil
\subfloat[Frame 5]{\includegraphics[width=0.16\textwidth]{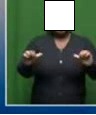}%
\label{fig:frame5}}
\hfil
\subfloat[Frame 6]{\includegraphics[width=0.16\textwidth]{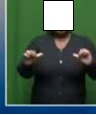}%
\label{fig:frame6}}
\caption{Representative video frames sampled uniformly from downloaded sign language videos.}
\label{fig:sample_video_frames}
\end{figure*}

Table~\ref{interpreter_methos_evaluation} shows the performance of our designed interpreter detection and extraction technique, reporting its true positive (TP), true negative (TN), false positives (FP), false negatives (FN), precision, recall, and F1 score, post manual corrections -- that is, after we manually labelled which videos contain Interpreters and which ones do not contain them. This is done so we can evaluate this approach analytically. Overall, our method shows promising results with 67$\%$, 82$\%$, and 74$\%$ precision, recall, and F1 score respectively. While our interpreter extraction method still requires improvements as shown by elevated false positive  -- predicting 27 interpreters where they do not exist, it significantly reduces the time to go through videos without interpreters as reported by high true negative. This is expected as many videos on SABC do not include interpretation. A further manual inspection shows that FN (i.e. videos missed but contain interpreters) may be due to a number of factors such as: poor video resolution, interpreter boxes existing temporarily in the video, and changes in the expected location and shape of the box, where the interpreter box is slightly up increasing its likelihood of being chopped into the bottom and top quadrants and the box shape taking a more rectangular shape than the expected square-like shape. FP, in contrast, occur mainly in videos featuring public interviews or protests, where several people appear on screen and additional faces may fall within the region of interest, leading to misclassification. We leave further investigations and improvements targeting trade-offs between TP, TN, FP, and FN for future works. With this, the sign language videos are extracted from the news video corpus, resulting in an example video as shown in Figure \ref{fig:sample_video_frames}. 

The parallel audio file can be trivially extracted from the $.webm$ or $.mp4$ audio stream, which we stored as a $.wav$ file using OpenCV~\footnote{https://pypi.org/project/opencv-python/} and ffmpeg~\footnote{https://www.ffmpeg.org}. Our objective was to align spoken language sentences with their corresponding sign language video sequences at sentence-level. However, the sign videos cropped are long videos that contain multiple sentences together with long audio files. To segment the audio streams, we used Automatic Speech Recognition (ASR) transcription to convert audio streams into text sequences for ease of tokenization using the NLTK toolkit. We expand on this in the next subsection -- Automatic Speech Recognition (ASR) Transcription. 
 
To segment sign language videos, on the other hand, we used a trivial motion detection algorithm that measures the average pixel-value difference between adjacent frames to discern motion and non-motion frames together with our proposed technique that uses gradient information (e.g direction) to locate probable segmentation points. We further defined N-breakpoints (where N is the number of sentences defined in the ASR phase during tokenization) to segment the videos using N lowest difference scores. We expand on each of these approaches subsequently.

\subsubsection{Automatic Speech Recognition (ASR) Transcription}
To transcribe the audio streams into text files, we used four state-of-the-art ASR models, namely, Whisper-large-v3 \cite{radford2022whisper}, WAV2VEC~\cite{baevski2020wav2vec}, Canary-Qwen-2.5B, and Granite-speech-3.3-8b. The first two models are primarily selected due to their widespread use in ASR research, while the last two are the top two models on an Open ASR leaderboard hosted on HuggingFace\footnote{https://huggingface.co/spaces/hf-audio/open\_asr\_leaderboard}, where adapted versions of Whisper and WAV2VEC are also found to be performing competitively. However, after using these four models for transcription, a manual inspection showed that the WAV2VEC transcriptions were not satisfactory, due to poor punctuation, all words being upper-cased, among other issues, and were left out for the rest of the experiments (examples provided in the dataset release repository). Due to not having annotated reference transcription, it was challenging to automatically verify the actual quality of all the transcriptions from the models, and select which model does better compared to others, other than manual inspection of a few samples. In this case, we resorted to only verifying using grammar checks and transcriptions' completeness through manual intervention. From this, we found Canary-2.5B to be generating quality transcriptions; however, it generated incomplete transcriptions, especially for long audio streams. Granite-speech-3.3-8b, on the other hand generated inconsistent transcriptions with multiple repetitions of phrases. The only model that generated quality transcriptions that required very little intervention for fixing was the Whisper-large-v3 model, which we generally adopt as the primary ASR transcriber for the remainder of this study. 

Despite its strong performance, Whisper-large-v3 exhibited several recurring issues, including -- misspelled proper names - especially of prominent public figures such as political leaders in South Africa (E.g., the Deputy president of South Africa \say{Paul Mashatile}, where Whisper-large-v3 generated - \say{Paul Machetele}, Canary-Qwen-2.5B generated \say{Paul Mashaetele} (close enough), and Granite-speech-3.3-8b generated \say{paul mashatele} (no capital letters to show this is a name of a person)); hallucination - the silence at the end of the audio stream provided noise to which Whisper-large-v3 transcribed as being audio signals; traces of multilingual training - where traces of the Chinese characters occur in transcriptions; repetition of sentences - Whisper-large-v3 generated duplicated transcriptions where interestingly, the last sentence would typically be a correction of the preceding errors resulting in the correct transcription (however still in the transcription); and incorrect sentence delimitation. However, these issues are more pronounced in other models, while they seldom occur in Whisper. Therefore, we used Whisper-large-v3 as the primary transcription model for this study. 

The resulting transcripts were manually post-edited to (i) standardize punctuation and sentence boundaries, (ii) remove incorrect or hallucinated segments, and (iii) translate code-switched words or non-English phrases into English where applicable. 

With clean transcription files available, we then used the number of sentences within each transcription file as a proxy to determine the number of times to segment the parallel video sequences. For example, if the parallel transcription file of a specific sign language video contains $N$ sentences, we immediately know we have to break the sign language video into $N$ shorter clips. We expand on the video segmentation algorithm subsequently.

\subsubsection{Sign Segmentation}
\begin{figure*}[!t]
\centering
\subfloat[Motion Detection Analysis]{\includegraphics[width=0.48\textwidth]{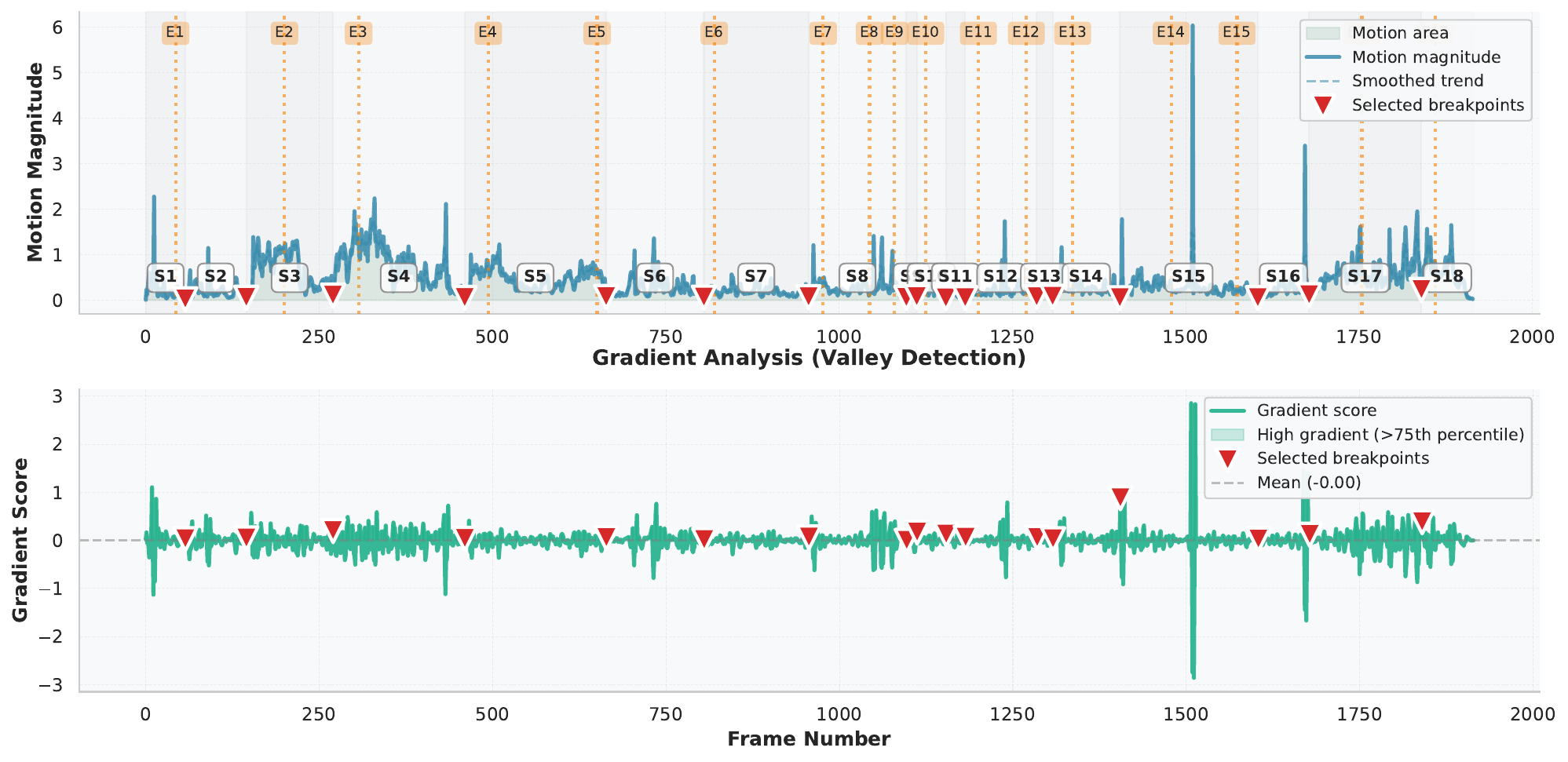}%
\label{fig:video split with grad and sent ratio}}
\hfil
\subfloat[Dataset Stratified Sampling]{\includegraphics[width=0.48\textwidth]{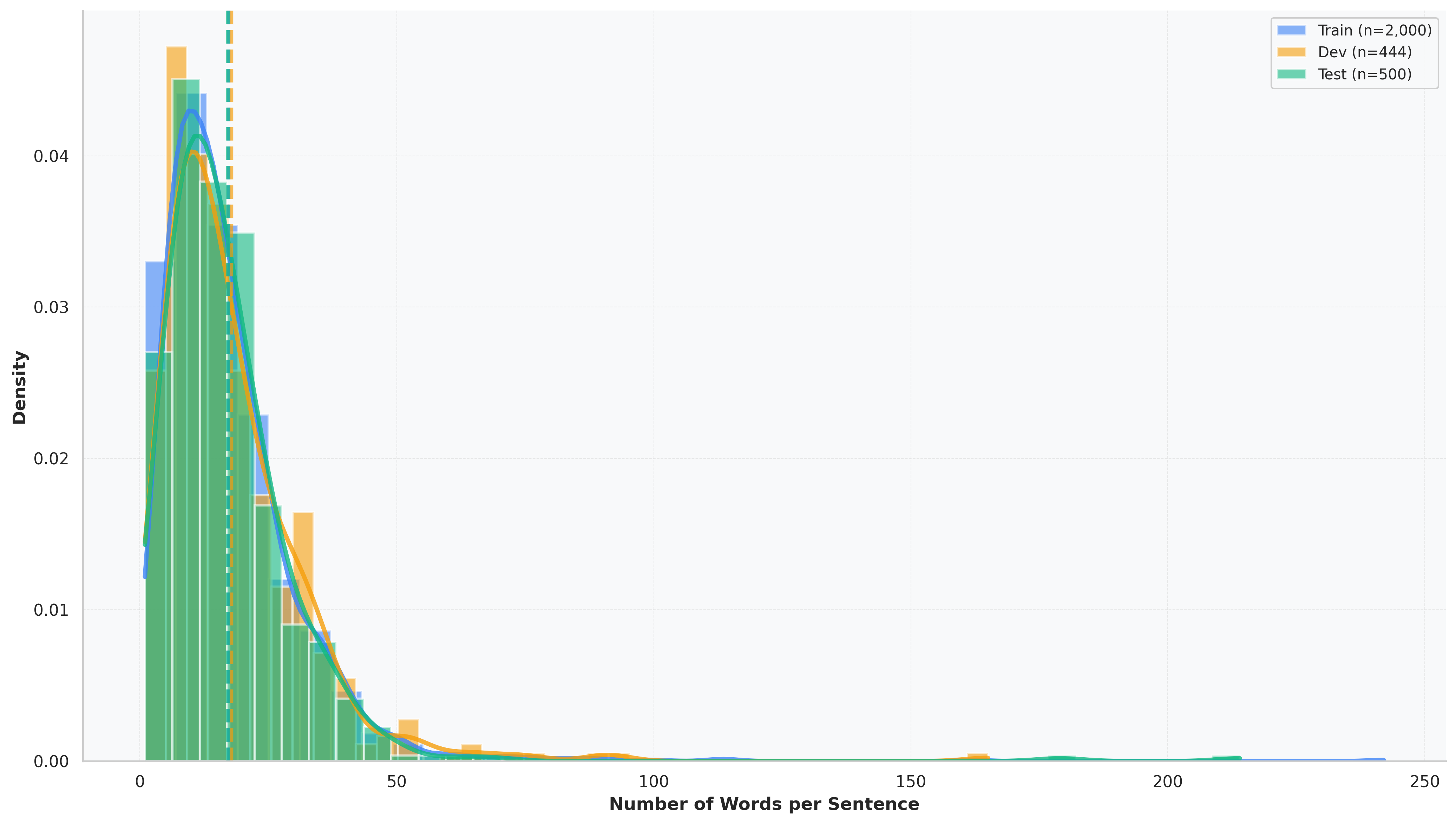}%
\label{fig:stratefied sampling train dev test}}
\caption{Left: A motion-halt detection graph with gradient-informed segmentation points with sentence-size ratios enforced (S(N)), and gradient detection without sentence ratios (E(N)) breaking points. N is the number of sentences determined during the audio transcription phase. Right: dataset splitting with stratified sampling to maintain similar sentence distributions in training, development, and testing.}
\label{fig:motion_and_sampling}
\end{figure*}

Many complex algorithms have been proposed for motion detection~\cite{noh2024decade} and sign language segmentation~\cite{moryossef2023linguistically}. However, in our study, we relied on a simple yet effective algorithm that subtracts the computed average frame value between two adjacent frames, as an indicator of motion occurrence or non-occurrence. Firstly, the frames are converted into greyscale images, and then projected into numpy array values as a preliminary pre-processing step. Then for each frame, a single average pixel value is calculated by calculating the average of all pixels to represent the frame. Then, by subtracting a predecessor-frame average value from a successor-frame average value (adjacent frames), we were able to get a value $\delta$ that highlights changes in pixel values (between frames $F_{t}$ and $F_{t+1}$ ), which was enough to infer the absence or presence of motion, given the static background of our sign language videos (see Figure \ref{fig:sabc_example} of the cropped section). That is, the only changes outside of random noise were the movements relative to signing in most videos. A low difference value indicates no movement, while large difference values show plausible changes in movement or movements. From the list of differences $\delta$, we identified the lowest $N$ values as our breaking points, where $N$ corresponds to the number of sentences in the equivalent transcription text file created in the ASR phase. That is, for $N$ sentences in the transcription file, we expected $N$ short video clips as segments. However, this did not perform well, as low points can appear within the occurrence of motion, leading to poorly cut videos. To address this, we used a gradient measure to find $N$ points with the highest descent rates (indicating a drop (change) in the motion difference leading to a plateau (low point), i.e., a stop or a no-motion indicator), which improved breaking quality. That is, we used gradients to find $N$ valleys of steepest descent. Empirical evaluation showed that a window size of 3 frames to measure gradient descent trajectories across the entire video showed improved results. Additionally, adjusting the video segmentation points based on the sentence-length distributions of the equivalent transcription files, further improved the quality of the gradient-based segmentation as shown in Figure \ref{fig:video split with grad and sent ratio}. This is done so that the longest sentence would have the longest video segment, while the shortest sentence in the transcription file will have the shortest video clip. Following this process, we chunked the long video stream into short segments that we aligned with the sentences in the equivalent transcribed file temporally (in order of appearance) to create our Maswao 'weakly' aligned SASL datasets. This process resulted in a corpus of 2944 video-to-text weakly aligned sentence pairs and a diverse vocabulary set (see Table~\ref{tab:Maswao_statistics}). More sophisticated motion detection and segmentation approaches that are heavily dependent on quality annotations such as prosody cues, or optical flow data~\cite{moryossef2023linguistically,rao2025sign} are left for future work.

\textbf{Maswao Training Details} To ensure uniform sentence distribution across our training, development, and testing datasets, we used stratified sampling, with the sentence lengths as classes for dataset splitting. From this, we split the data into 2000, 444, 500 aligned video-to-text data for training, development, and testing as shown in Figure \ref{fig:stratefied sampling train dev test} (following the 80:20 split approach), respectively. This is to ensure that training and evaluation are done with the same dataset distribution. 

\subsection{Sign-to-text Translation Baselines}
We validated the quality of the weakly aligned Maswao dataset through a series of training, development and testing experiments using three state-of-the-art sign-to-text translation (STT) baselines: Neural SLT (NSLT)~\cite{camgoz2018neural}, Joint SLT~\cite{camgoz2020sign}, and Gloss-Free End-to-End SLT (GloFE)~\cite{lin-etal-2023-gloss}. NSLT follows an encoder–decoder sequence-to-sequence design based on recurrent neural networks~\cite{luong2015effective}, combined with either \cite{luong2015effective} or \cite{bahdanau2014neural} attention mechanisms. In contrast, Joint SLT and GloFE adopt Transformer backbones. This selection enables a direct comparison between recurrent sequence-to-sequence models and modern Transformer-based architectures. The baselines also differ in input modality. NSLT and Joint SLT operate on RGB image input, using AlexNet~\cite{krizhevsky2012imagenet} as the feature extractor, while GloFE uses skeletal pose features estimated with MMPose~\cite{MMPose2020}. Evaluating both image-based and pose-based inputs provides a more comprehensive assessment of the dataset and the effects of representation choices. For evaluation, we report BLEU-1 to BLEU-4 (Bilingual Evaluation Understudy) and ROUGE-L (Recall-Oriented Understudy for Gisting Evaluation), consistent with common practice in STT benchmarking~\cite{yin2020better,wang2024generalizations, lopez2025multitask}.

\begin{figure*}[!t]
\centering
\subfloat[BLEU-1]{\includegraphics[width=0.24\textwidth]{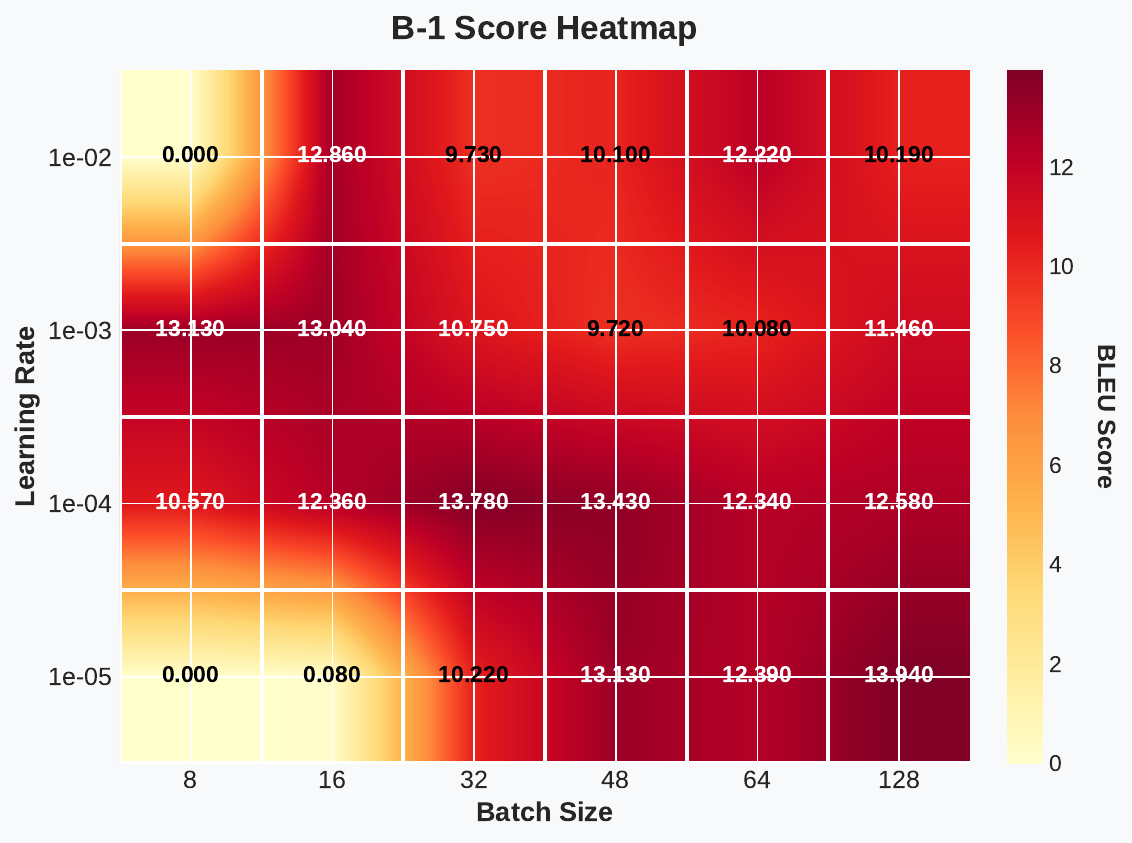}%
\label{fig:bleu1}}
\hfil
\subfloat[BLEU-2]{\includegraphics[width=0.24\textwidth]{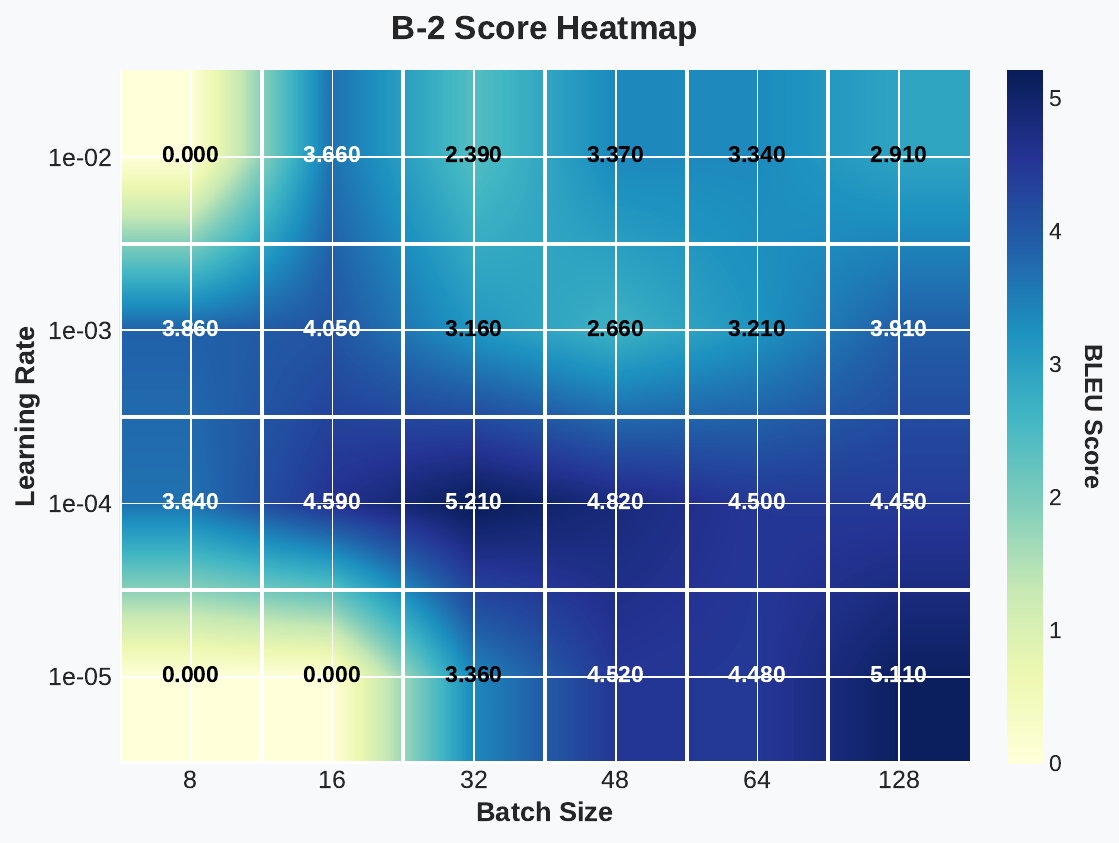}%
\label{fig:bleu2}}
\hfil
\subfloat[BLEU-3]{\includegraphics[width=0.24\textwidth]{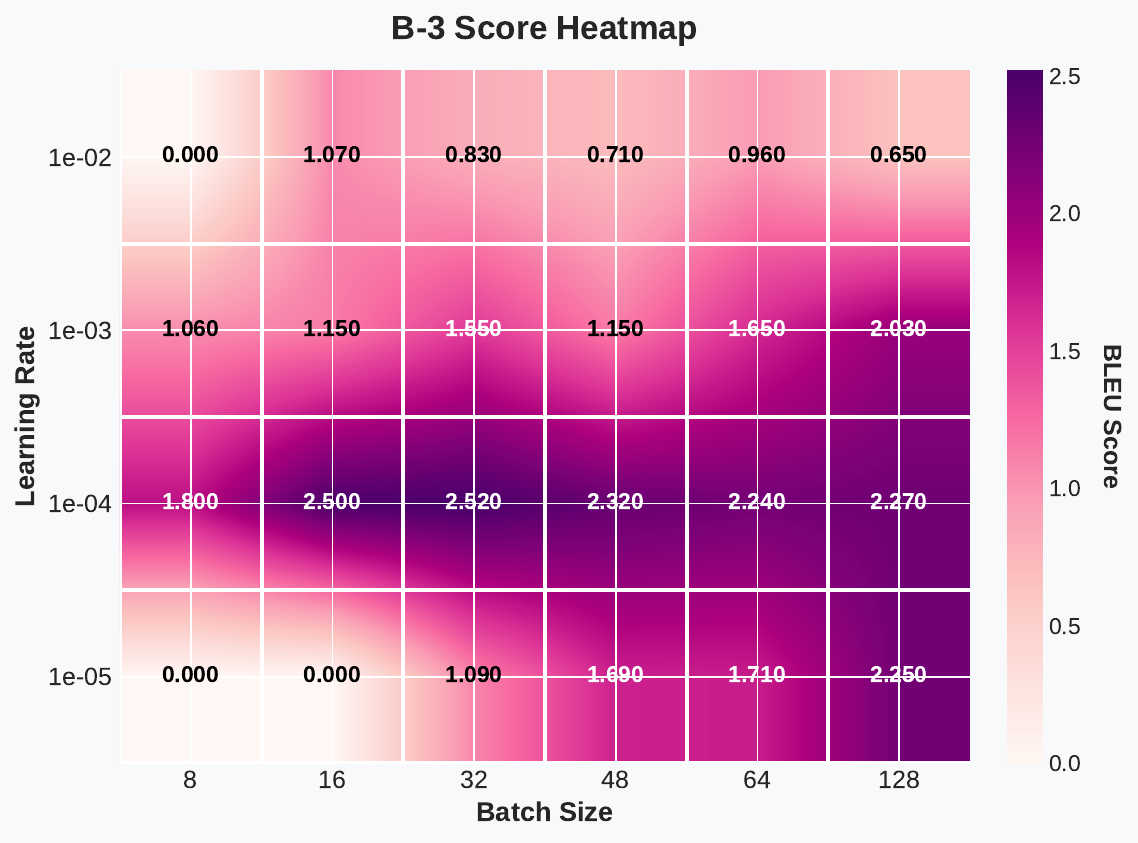}%
\label{fig:bleu3}}
\hfil
\subfloat[BLEU-4]{\includegraphics[width=0.24\textwidth]{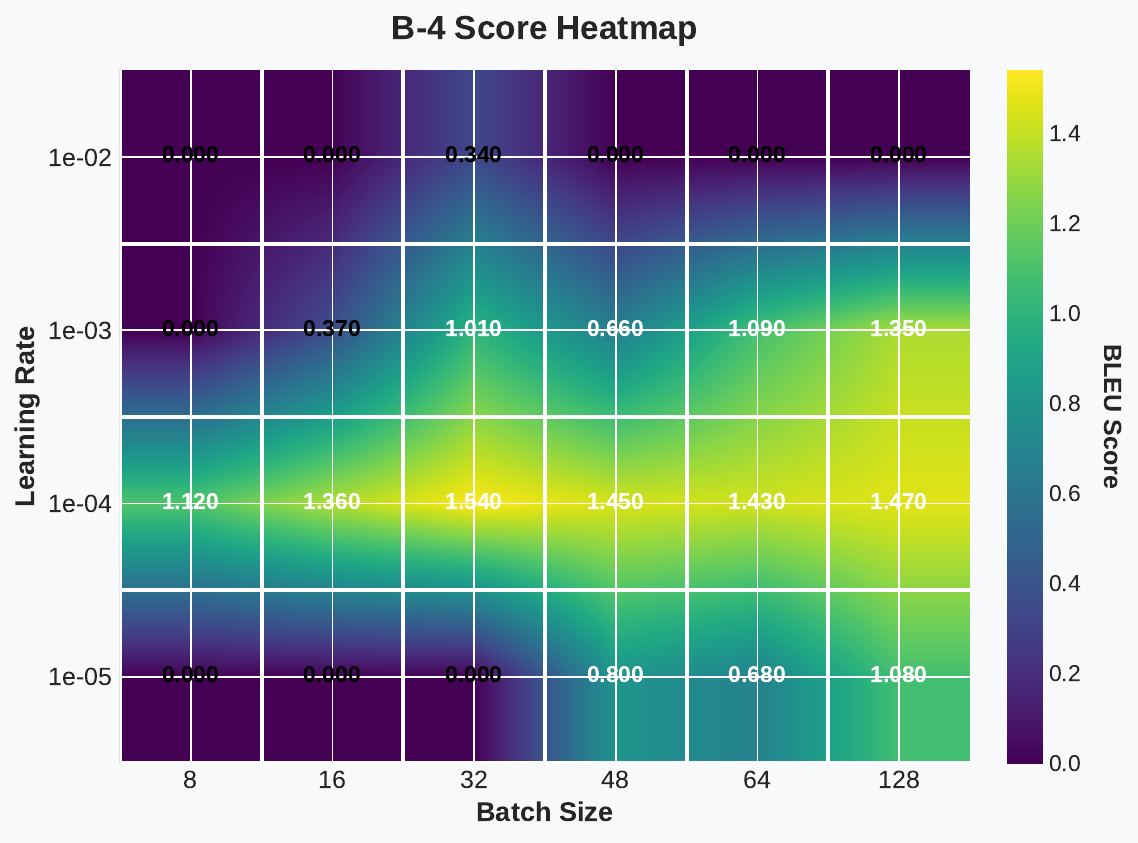}%
\label{fig:bleu4}}
\caption{BLEU Score Heatmaps: Impact of Batch Size and Learning Rate. Higher scores indicate better translation quality.}
\label{fig:bleu_heatmaps}
\end{figure*}

\textbf{Baseline Models Hyperparameters} We experimented using mostly default hyper-parameter setups (where applicable) for training, development, and testing as originally proposed in respective works of our baseline models. These include using the same network architectures, such as the number of layers, batch size (32), optimizing algorithms (Adam optimizer), evaluation metrics -- BLEU 1-4, with an addition of ChrF and ROUGE-L. Although Transformers showed improved translation performance over recurrent models, we further explored the hyper-parameter search by comparing different batch sizes in the set $\{16, 32, 48, 64, 128\}$ with various learning rates $\{0.01, ..., 0.00001 \}$. This highlighted that a batch size of 32 and a learning rate of $0.0001$ illustrated improved performance as shown in Figure \ref{fig:bleu_heatmaps}, which we used to train our model for much longer epochs to establish our best performing model.

\section{Results}\label{sec2}
In this Section, we present quantitative evaluation of the Maswao corpus and summarize the baseline performance of sign-to-text translations. 

\subsection{Sign-to-Text Translation}

\begin{table*}[t]
\centering
\small
\setlength{\tabcolsep}{4pt}
\resizebox{\textwidth}{!}{%
\begin{tabular}{l|cccccc|cccccc}
\hline
\textbf{Model} 
& \multicolumn{6}{c|}{\textbf{DEV}} 
& \multicolumn{6}{c}{\textbf{Test}} \\
\hline

& \textbf{B1} & \textbf{B2} & \textbf{B3} & \textbf{B4} & \textbf{CHRF} & \textbf{ROUGE}
& \textbf{B1} & \textbf{B2} & \textbf{B3} & \textbf{B4} & \textbf{CHRF} & \textbf{ROUGE} \\
\hline

NSLT-gru-luo      
& 3.73 & 0.66 & 0.20 & 0.0 & 10.16 & 3.58  
& 3.81 & 0.71 & 0.0 & 0.0 & 10.11 & 3.3 \\


Joint SLT-1024
& 13.78 & 5.26 & 2.71 & 1.74 & 17.86 & 10.81 
& 13.78 & 5.21 & 2.52 & 1.54 & 18.09 & 10.51 \\

Joint SLT-4096
 & 12.28 & 5.05 & 3 & $\textbf{2.17}$ & 18.36 & 9.93 
& 11.63 &  4.18 & 2.19 & 1.45 & 17.95 & 8.80 \\

GloFE     
& 12.66 & 3.68 & 1.79 & 1.18 & -- & -- 
& 12.37 & 3.39 & 1.35 & 0.79 & -- & -- \\

\hline
\textbf{Our SLT} 
& 12.93 & 4.92 & 2.68 & 1.79 & 18.33 & 10.03 
& 12.86 & 4.67 & 2.28 & 1.46 & 18.25 & 9.64 \\
\hline
\end{tabular}
}
\caption{Translation performance on the Maswao DEV and Test sets. We report BLEU-1 to BLEU-4 (B1--B4), CHRF, and ROUGE scores for NSLT, GloFE, and Joint SLT models, and our proposed STT model. We ran each model five times with a different random seed and reported the average results.}
\label{tab:slt_results}
\end{table*}

Table \ref{tab:slt_results} shows a performance comparison of recurrent-based neural networks (LSTM vs GRU with \cite{luong2015effective} and \cite{bahdanau2014neural} attention) trained with RGB features extracted using AlexNet CNN model -- (NSLT), a Transformer-based model with RGB features extracted using AlexNet (using output features vector dimensions of 1024 and 4096) -- Joint SLT, and a Transformer-based model trained with skeletal pose features extracted using a pose estimation tool, MMPose -- GloFe, covering diverse modeling techniques and features extraction methodologies. 

From these results, it is apparent that Joint SLT outperforms recurrent-based networks and skeletal-based models across all metrics, BLEU 1-4, ChrF, and ROUGE-L. This is in line with observations highlighted in existing literature, that RGB-based feature extractions outperforms skeletal based feature extraction techniques \cite{ko2019neural,orbay2020neural}. This behaviour is compounded when we use larger dimension size (4096) features of the AlexNet model. We hypothesize that the performance degradation of skeletal estimation may be affected by the quality of the sign language videos, which warrants further investigation as studies showed that the quality of videos negatively impacted the estimation of subtle but crucial features, such the fingers \cite{de2020sign}. Compounding the poor performance of competing models NSLT and GloFE, is that, due to a lack of matching resources used in their respective studies, their techniques, such as the contrastive learning (which required linguistic insights such as parts of speech tags) in GloFE, and supervision signals from glosses in NSLT where omitted from experiments, which may have negatively impacted training performance. Consequently, we adopted Joint SLT, with optimal training parameters (batch size=32, learning rate=$1e^{-4}$), with longer early stopping patience=$50$ to train for longer times (epochs = 616) to establish our model -- \textbf{Our STT}. Our STT translation model established a benchmark for our automatically collected and prepared, weakly aligned Maswao dataset, curving new paths for future works in sign language translation for the South African Sign Language (SASL) community. Moreover, this study pioneers works in automatic collection and preparation (i.e., alignment of sentences to parallel video chunks in long videos) of low-resourced sign language datasets with similar resource-availability characteristics.    

Comparative analysis of recurrent architectures demonstrates that an GRU-based recurrent models integrated with Luong attention \cite{luong2015effective} achieves superior performance compared to GRU with Bahdanau attention \cite{bahdanau2014neural} or LSTM configurations using either Luong attention \cite{luong2015effective} or Bahdanau attention \cite{bahdanau2014neural}. Nevertheless, despite these findings, all recurrent-based models underperform relative to Transformer architectures and additionally incur substantially higher training times. 

\subsubsection{Additional Results}
To complement lexical-overlap metrics, which penalise valid paraphrases and say little about meaning, we evaluated the best-performing sign-to-text model with BERTScore and MetricX. BERTScore compares candidate and reference sentences using contextual token embeddings from a pretrained language model, computing precision, recall and F1-score from soft token matches, so it rewards semantic similarity rather than exact word overlap~\cite{zhang2019bertscore}. MetricX-24 is a learned, regression-based translation quality metric built on mT5 and trained on human quality ratings, which predicts an error-style score where lower values indicate better translations~\cite{juraska-etal-2024-metricx}. Together, these metrics matter for our weakly aligned Maswao dataset because a sign video can be translated in several valid ways while only a single reference is available, so generated outputs may be semantically related to the reference without being an exact match. As shown in Table~\ref{tab:bertscore_metricx}, our best-performing model obtains a BERTScore of 12.20 (P), 14.60 (R) and 13.23 (F1), and a MetricX error score of 21.85. These values indicate limited semantic fidelity and substantial translation errors, and the higher recall than precision suggests that the model captures part of the reference meaning but also produces additional or irrelevant content. Such results are expected given the difficulty of open-domain sign-to-text translation, which may be compounded by the diverse vocabulary of Maswao and by signer variability. Importantly, the objective of this work was not to develop the best possible model but to propose a dataset; the model serves as a reproducible baseline showing that Maswao can be used to train and evaluate sign-to-text systems, and the remaining headroom provides a clear benchmark for future work. To better understand these results, we further analyse the outputs of our best model through a qualitative comparison between the generated and reference translations, reported in the following section.

\begin{table}[t]
\centering
\begin{tabular}{lcccc}
\hline
\textbf{Model} &
\multicolumn{3}{c}{\textbf{BERTScore}} &
\textbf{MetricX} \\
\cline{2-4}
& \textbf{P} & \textbf{R} & \textbf{F1} & \textbf{Score} \\
\hline
Our Model & 12.20 & 14.60 & 13.23 & 21.85 \\
\hline
\end{tabular}
\caption{Complementary evaluation of the best-performing
sign-to-text model using BERTScore and MetricX.}
\label{tab:bertscore_metricx}
\end{table}

\section{Qualitative Results}\label{sec2.2}
In this section, we share qualitative results of our best-performing sign language translation model. Table \ref{tab:qualitative results} shows four example translations of our model compared to the ground truth (GT) results.

\begin{table}[!t]
\centering
\renewcommand{\arraystretch}{1.2}
\setlength{\tabcolsep}{4pt}
\begin{tabular}{|c|p{0.7\columnwidth}|} 
\hline
\textbf{GT} & And it was incumbent upon us to set up policies. \\
\textbf{Our STT} & This is but we do that we had the \textcolor{green}{policies} which we had to \textcolor{green}{set up} regarding the language. \\
\hline\hline
\textbf{GT} & Do you see police visibility? \\
\textbf{Our STT} & They are still shooting on the \textcolor{green}{police} around here? \\
\hline\hline
\textbf{GT} & And it is spoken by blacks in not only so‑called colored townships. \\
\textbf{Our STT} & That's the Afrikaans language has been \textcolor{green}{spoken} as you just heard. \\
\hline\hline
\textbf{GT} & The 6 am Central African Time you're watching SABC News live from Johannesburg. \\
\textbf{Our STT} & You're \textcolor{green}{watching SABC News} live from \textcolor{green}{Johannesburg} in your headlines this hour. \\
\hline
\end{tabular}
\caption{Translations from our neural network model using Maswao datasets. GT = Ground Truth, STT = System Output. Highlighted words indicate correct matches.}\label{tab:qualitative results}
\end{table}

From this, we can observe that our model is able to generate sound, coherent, and structurally correct sentences. Although these sentences do not generally match the semantics of the ground truth labels, our model is able to generate keywords (highlighted in green) appearing in the ground truth. Furthermore, the models were able to capture context (row 3) as Afrikaans is a language spoken by the colored communities of South Africa. This probably means that somewhere in the embedding space, the representations of the word colored and Afrikaans were collocated. However, this warrants further investigation. We hypothesize that with increased training examples and higher quality videos, better modeling, improved segmentations, results on the Maswao datasets would lead to more accurate translations.   




\section{Conclusion}\label{sec13}
The Deaf and Hard-of-hearing communities have been sidelined from many societal benefits, lending perpetuated marginalization of the communities and reduced access to important socio-economic activities. While advancements in sign-to-text and text-to-sign research have been used to address this from a communication-facilitation perspective, issues of unequal research efforts arise, where only a few languages reap the benefits of important strides. To address this disparity, we propose a reproducible framework that details the automatic collection and preparation of digital news content of sign languages, using the South African Sign Language (SASL) as a case study. Following this, we create sequences of weakly aligned sign language short videos to spoken language text sentences, using simple yet effective techniques such as automated speech recognition, non-motion detection with gradient descent, and sentence-length distributions of the accompanying transcription file weighting. The quality of our datasets is then evaluated using MMPose, AlexNet features, recurrent neural networks and Transformer architectures, highlighting the efficacy of our automated approach to dataset collection and preparation. Concretely, our results show that Transformer-based modeling with RGB features extracted using AlexNet, outperforms recurrent neural networks with the same features, and Transformer models with skeletal pose estimation features, on our proposed Maswao dataset.  

We hope to see future works exploring other pose estimation techniques, such as OpenPose and MediaPipe, for improved skeletal pose features. Additionally, we hope to see improvements in video segmentation for better alignments of sentences. Furthermore, with this complex task and dataset, we hope to see improvements in model complexity to capture the intricacies in sign language translation. Moreover, Text-to-sign (TTS) has been considered a much more challenging task compared sign-to-text. As such, our future works will build on this to include experiments for TTS for building end-to-end sign language translation systems for Deaf communities. Finally, our next milestone is to scale the framework we proposed here to include multiple South African languages, with rigorous validation steps across phases to improve dataset quality.

\section{Limitations}
Due to financial and resource constraints, this study did not annotate ground truth video segments to evaluate the automatic segmentation technique proposed to create our weakly aligned sentence-level Maswao dataset. However, through a manual inspection of video samples, it can be observed that our proposed technique segments videos exactly where the movement of the signer is reaching a stop. Notably, isolated lexical items (e.g., the word \say{Interesting}), which appeared as sentence number 93 from a total of 94 sentences, from the video with YouTube id \say{1-UpxWii8Yg} was segmented correctly as shown in Segment 93 of folder \say{1-UpxWii8Yg} of our released dataset. This provides evidence of the effectiveness of the proposed gradient-descent and sentence-length–based segmentation strategy. 

Similarly, the study did not have ground truth or labeled transcriptions to compare ASR techniques. As future work, it would be valuable to produce ground-truth transcriptions for a representative subset of the corpus and to investigate automatic transcriber selection using recent SOTA large language models (e.g., ChatGPT, LLaMA, and Gemini). In addition, exploring ranking-based selection frameworks may further improve the reliability and consistency of transcription model choice.

Although the adopted Automatic Speech Recognition (ASR) system produced high-quality transcriptions overall, the generated transcripts were not systematically validated by independent human translators. This is particularly relevant in cases where news presenters or interviewed persons code-switched into local languages, which can reduce ASR accuracy. Instead in this work, the authors triaged through the transcriptions to manually fix identified errors, similar to \cite{tanzer2025youtube}. While this approach enabled the construction of the dataset, a comprehensive human validation of the transcriptions would further improve dataset quality, quantify the impact of transcription errors, and help identify potential sources of noise that remain undetected. 

Owing to YouTube's licensing terms and our ethical responsibility to protect the privacy of the sign language interpreters whose faces appear in the videos, we are unable to publicly redistribute the original sign language video clips included in the Maswao dataset. Instead, we release the corresponding YouTube video identifiers (IDs) and the accompanying sentence-level text annotations, allowing researchers to reconstruct the dataset from the original publicly available sources while respecting both platform licensing requirements and the privacy of the individuals depicted.

\bibliographystyle{unsrtnat}
\bibliography{references}  






\end{document}